\pdfoutput=1
\documentclass[sigconf,nonacm]{acmart}

\usepackage{array}
\usepackage{microtype}

\newcommand{\fnl}{\textsc{Free}}
\newcommand{\cnl}{\textsc{Constr}}
\newcommand{\json}{\textsc{Json}}
\newcommand{\triples}{\textsc{Triples}}
\newcommand{\kv}{\textsc{Kv}}
\newcommand{\qarec}{QA recall}

\title{Faithful, Not Corrective: Model Capability Governs Message-Format Effects in Multi-Hop Agent Relays}

\author{Sicheng Zeng}
\email{1244537181@qq.com}
\affiliation{%
  \institution{Independent Researcher}
  \country{}
}

\begin{document}

\begin{abstract}
When LLM agents hand information to one another, does the message format matter?
Two literatures disagree: format-optimization work reports that structured messages cut cost without hurting accuracy, while format-restriction studies find that imposing structure degrades generation.
Neither line has measured what happens when messages traverse \emph{multiple} hops, where copy fidelity, rather than one-shot generation quality, dominates.
We introduce a controlled relay testbed in which briefs of twelve programmatic atomic facts are re-encoded hop by hop in five formats (free natural language, precision-instructed NL, JSON, triples, key--value) over six hops, scored against programmatic ground truth by a fixed strong grader, across two relay-capability tiers, a cognitive-load condition, and a paired-fork error injection.
We find that (i) a strong relay is nearly lossless for every format (hop-6 \qarec{} $\geq 0.973$), with residual loss concentrated at the first encoding step; (ii) per-hop cognitive load raises generation cost by 24--53\% while fidelity changes stay within $\pm 1.8$ points; (iii) under a weak 1.5B relay, the across-format dispersion of hop-6 recall grows by a factor of $8.7$ (CI $5.3$--$15.5$), driven by an encode--drift trade-off that flips the format ranking in transit; and (iv) once an injected error is present, every format propagates it faithfully (surface persistence 83--100\%) and no format cascades collateral damage onto neighboring facts.
Structure buys a faithful, error-localizing channel, not an error-correcting code.
\end{abstract}

\ccsdesc[500]{Computing methodologies~Multi-agent systems}
\ccsdesc[300]{Computing methodologies~Cooperation and coordination}
\ccsdesc[100]{Computing methodologies~Natural language processing}

\keywords{multi-agent LLM systems, agent communication, message formats, information relay, fidelity, capability stratification, evaluation methodology}

\maketitle

\section{Introduction}

Multi-agent LLM systems are, at bottom, relay systems: planners, executors, reviewers, and orchestrators re-encode task-critical information at every boundary \cite{hong2023metagpt,qian2023chatdev,du2023debate}.
Three properties of how these systems are deployed turn the reliability of each handoff into a design question rather than an implementation detail.
Cost pressure is moving small models into boundary roles \cite{narayan2025minions,zywot2026small}; cross-vendor pipelines are black-box, so the text message is the only interface a designer can actually specify; and failure taxonomies attribute over a third of multi-agent failures to inter-agent misalignment rather than to base-model capability \cite{cemri2025mast}.
What gets lost at each handoff, and what the message \emph{format} does to that loss, is therefore a load-bearing decision for distributed agentic systems.

Two adjacent literatures answer it in opposite directions.
Format-optimization work shows that structured, non-natural-language formats preserve accuracy at much lower token cost \cite{chen2024autoform}, and protocol guidelines recommend schema-constrained messages to prevent semantic drift \cite{wibowo2025safeagents}.
Format-restriction studies find the opposite: rigid schemas measurably degrade reasoning \cite{tam2024speakfreely}, and natural-language tool interfaces outperform structured schemas \cite{johnson2025nltools}.
Meanwhile, a distortion literature documents that information passed through chains of LLM generations degrades progressively, the ``broken telephone'' effect \cite{mohamed2025telephone,perez2024telephone,ghafouri2026lost}, with delegation accuracy collapsing from 90.7\% to 22.5\% over five stages in one study \cite{ao2026reliability}, and theory bounding the accumulation of relay error \cite{fan2026martingale}.

These lines have never met, and the reason is structural rather than incidental.
Distortion studies fix the format and vary the chain; format studies vary the format over shallow, fixed-depth exchanges, without per-hop fidelity measurement, because their benchmarks are built for one-shot generation rather than for relay.
The missing cell matters because multi-hop relay changes which mechanism dominates.
At one hop the question is whether structure helps a model \emph{generate}; across hops it is whether structure helps a model \emph{re-encode} content it did not author.
The sign of the format effect in the relay regime is genuinely unknown, and it is precisely the regime that cost-driven deployment now occupies.

We resolve it with a controlled testbed in which everything that threatens validity is pinned down.
Ground truth is \emph{programmatic}: briefs are rendered from twelve synthetic atomic facts, so fidelity is measured against the generator, never another model's output.
Each relay agent sees only the previous hop's message and must re-issue the complete handoff in the same format.
Fidelity is scored two ways: a paraphrase-tolerant \qarec{} (a fixed strong grader answers twelve closed-form questions, exact-matched programmatically) and a judge-free verbatim string recall; the grader is held fixed across all conditions so relay capability is never conflated with grader capability.
We cross five formats with six hops, two relay-capability tiers (a strong API model versus a 1.5B local model), a cognitive-load condition at the strong tier, and a paired-fork causal injection at the weak tier.

\paragraph{Contributions.}
(1)~We complete the format~$\times$~hops~$\times$~capability design space with the first format-controlled relay study: for a strong relay the telephone-game collapse does not occur (\qarec{} $\geq 0.973$ after six hops), and for all but free NL the residual loss concentrates at the first encoding step (\S\ref{sec:strong}).
(2)~At the strong tier, per-hop cognitive load raises generation cost by 24--53\% while any fidelity effect is bounded within $\pm 1.8$ points (\S\ref{sec:load}).
(3)~We show the format effect is a \emph{capability regime}: under the weak relay the across-format dispersion of hop-6 recall grows $8.7\times$ (CI $5.3$--$15.5$; range $2.3\to20.5$ points), with paired weak$-$strong hop-6 differences of 18--37 points.
The effect decomposes into an encoding toll paid by rigid formats versus drift resistance tied specifically to the fixed-key JSON schema, which flips the format ranking in transit (\S\ref{sec:weak}).
(4)~With a paired-fork injection we causally adjudicate the ``schemas prevent drift'' folklore: once an injected error is present, every format propagates it faithfully (83--100\% surface persistence at the final hop), and no format exhibits detectable collateral cascade.
Structure is a faithful, error-localizing channel, not an error-correcting code (\S\ref{sec:causal}).

\section{Related Work}

\paragraph{Iterated distortion in LLM chains.}
\citet{mohamed2025telephone} establish that iterative LLM generation distorts information in translation chains; \citet{perez2024telephone} characterize attractor dynamics over repeated transmission, and \citet{ghafouri2026lost} track which social information survives AI--AI relay.
Closest in machinery, \citet{fan2026martingale} bound error accumulation across sequential tool calls, and \citet{ao2026reliability} derive mutual-information limits for delegated multi-stage pipelines and observe accuracy collapsing within five stages.
All of these hold the message format fixed; none treats format as the experimental variable, and most rely on translation or social-content proxies rather than task-grounded facts with programmatic ground truth.

\paragraph{Format effects at a single exchange.}
\citet{chen2024autoform} show agents can adopt compact non-NL formats with comparable accuracy at up to 72.7\% fewer tokens; structured inter-agent protocols are standard in deployed frameworks \cite{wang2025talkhier,marro2024agora,ehtesham2025protocolsurvey}.
The opposing pole reports that format \emph{restrictions} degrade reasoning \cite{tam2024speakfreely}, a contested finding whose size depends on prompt engineering, and that natural-language interfaces beat structured schemas for tool use \cite{johnson2025nltools}; prompt-format sensitivity is itself large \cite{sclar2023format}.
These single-exchange results cannot be extrapolated to relays, where per-hop copy fidelity, not generation quality, dominates, the regime we measure.

\paragraph{Multi-agent communication and its failures.}
Topology shapes how errors and insights propagate \cite{shen2025eib}, seeded errors can solidify into false consensus \cite{xie2026spark}, and failure taxonomies place inter-agent misalignment among the leading causes of system failure \cite{cemri2025mast}.
Capability is an emerging moderator: architecture--task fit governs when coordination helps \cite{kim2025scaling}, and compute-matched comparisons reverse headline multi-agent gains \cite{tran2026equal}.
Orthogonal work replaces the text channel with activations, latents, or KV caches \cite{ramesh2025activations,du2025interlat,shi2025kvcomm}; such channels require white-box access to every model, which cross-vendor pipelines do not provide, so we characterize the text channel that black-box pipelines actually use.
Our experimental hygiene follows recent methodological guidance for collective-LLM experiments \cite{zhou2025pimmur}.

\section{The Relay Testbed}
\label{sec:testbed}

A relay measurement is only informative if nothing else can masquerade as a format effect, so the testbed is organized around the threats it must neutralize rather than around components.
Five threats dominate, and each design element below exists because of one of them:
(1)~\emph{ground truth by proxy}, where scoring against another model's output confuses fidelity with inter-model agreement (\S\ref{sec:corpus});
(2)~\emph{paraphrase}, which a verbatim metric misreads as loss (\S\ref{sec:measurement});
(3)~\emph{grader--relay conflation}, where a grader whose capability varies with the relay tier confuses reading with relaying (\S\ref{sec:tiers});
(4)~\emph{cost confounds}, where hidden reasoning tokens distort token comparisons (\S\ref{sec:measurement});
and (5)~\emph{causal ambiguity}, where the folklore claim ``structure prevents drift'' bundles resisting change with repairing corruption (\S\ref{sec:inject}).

\subsection{Corpus with programmatic ground truth}
\label{sec:corpus}
Each item is a task-handoff brief rendered by deterministic templates from twelve atomic facts in three domains (software migration, incident response, logistics): person--role bindings, quantities, constraints/deadlines, and dependencies/prohibitions.
Ground truth is the generating fact tuple itself, never any model output, and a self-check verifies that every fact's surface form appears verbatim in the hop-0 brief (string recall $=1.0$ by construction).
Synthetic values also block a grader from answering out of world knowledge.

\subsection{Relay protocol}
Hop~1 receives the original brief and re-issues the complete handoff in the assigned format; hop~$i{>}1$ receives \emph{only} hop~$(i{-}1)$'s message.
Relay temperature is $0.7$ throughout.
In the \emph{process-relay} condition, each hop first answers, under an \textsc{Analysis} heading, which element of the handoff is most time-critical and why, before re-issuing the handoff; the analysis section is stripped programmatically before forwarding.

\subsection{Formats}
\label{sec:formats}
Five format specifications span the structure spectrum:
\fnl{} (free prose, no constraints);
\cnl{} (natural language plus a precision instruction: convey every fact, keep numbers/names/codes verbatim);
\json{} (a fixed six-key schema: title, people, quantities, constraints, dependencies, prohibitions);
\triples{} (subject\,$|$\,predicate\,$|$\,object lines);
\kv{} (flat snake\_case key--value lines).
JSON messages are checked for syntactic validity at every hop; invalid messages (1\% under the weak relay, none under the strong) are passed downstream and scored as-is.

\subsection{Measurement}
\label{sec:measurement}
\textbf{\qarec{}} (primary): a fixed grader, the strong model at temperature~$0$, reads only the hop message and answers the twelve closed-form questions in JSON; answers are exact-matched programmatically against ground truth.
The grader never sees the ground truth or the original brief, and grader parse failures trigger one retry and are otherwise excluded (parse-fail rate $0.0\%$ in all grids).
\textbf{String recall} (secondary, judge-free): normalized verbatim containment of each fact's surface form.
String recall under-counts formats that legitimately re-render surface forms while preserving content: in the strong grid the hop-6 QA$-$string gap is $+8.3$ points for \kv{} and $+4.0$ for \triples{}, which is why a paraphrase-tolerant primary metric is required.
In the weak grid the gap reverses sign for three formats (QA 3--5 points \emph{below} string recall), consistent with degraded messages preserving surface strings while corrupting relations; we verify all weak-tier orderings under both metrics (Appendix~\ref{app:mimo}).
Token accounting excludes hidden reasoning tokens, which would otherwise inflate message-length comparisons by up to $4\times$ (Appendix~\ref{app:pareto}).

\subsection{Capability tiers and grader separation}
\label{sec:tiers}
Strong-tier grids use a strong API model as the relay; the weak-tier grid uses Qwen2.5-1.5B-Instruct \cite{qwen25} (4-bit quantized, served locally).
Crucially, the \emph{grader is the strong model in every condition}, so weak-relay distortion is never conflated with weak grading.
Two facts argue against a grader artifact: first, the judge-free string metric independently reproduces the weak-tier separation ($22.2$-point hop-6 spread) and the flat-\json{} signature (Appendix~\ref{app:mimo}); second, on well-formed messages the grader reads every format at $\geq 0.973$ (\S\ref{sec:strong}), bounding format-specific reading loss at $\approx$2.7 points.

\subsection{Causal error injection}
\label{sec:inject}
To separate ``structure prevents change'' from ``structure repairs errors,'' we use a paired fork: hops 1--3 run normally; at hop~3's output we programmatically replace one owner name (all surface occurrences) with a plausible name absent from the item; the chain then forks into clean and injected arms for hops 4--6.
Both arms share the identical prefix, so any downstream difference is the causal effect of the single injected error.
Injection is only possible when the target name is still present at hop~3; this held in 164 of 200 chains (82\%), with per-format opportunity rates mirroring overall fidelity.
All injection analyses condition on successful injection.
We track \emph{persistence} (does the wrong value survive to hop~6?) and \emph{containment} (paired difference in recall over the eleven uninjected facts).

\begin{table}[t]
\centering
\small
\begin{tabular}{@{}llll@{}}
\toprule
Grid & Relay & Design & Scored \\
\midrule
Strong & strong & $100 \times 5 \times 6$, pure & 3{,}000 \\
Load & strong & $60 \times 5 \times 6$, process & 1{,}800 \\
Weak & weak & $50 \times 5 \times 6$, pure & 1{,}500 \\
Inject & weak & $40 \times 5$, fork@3 & 1{,}800 \\
\bottomrule
\end{tabular}
\caption{Design matrix (items $\times$ formats $\times$ hops; ``Scored'' counts graded messages). The Inject grid scores $40\times5$ chains $\times$ 9 messages each (3 shared prefix + 3 clean-arm + 3 injected-arm hops). Every grid is graded by the same fixed strong grader at $t{=}0$.}
\label{tab:design}
\end{table}

\section{Experiments}

Table~\ref{tab:design} summarizes the four grids.
All uncertainty is reported as percentile-bootstrap 95\% CIs over items (2{,}000--4{,}000 resamples; pointwise unless stated); proportions carry Wilson 95\% CIs; one ``point'' is $0.01$ recall.
Within each analysis family we report a Bonferroni-adjusted simultaneous bound where noted (load, drift advantages) or flag comparisons as uncorrected (containment).
Decay slopes fit a line to logit-transformed per-hop mean recall over hops 1--6 (means clipped to $[10^{-3}, 1-10^{-3}]$), with CIs from a bootstrap over items; raw hop-1$\to$6 point losses are reported alongside every slope.

\subsection{A strong relay barely distorts, and the loss lives at the encoding step}
\label{sec:strong}

\begin{figure*}[t]
\centering
\includegraphics[width=\textwidth]{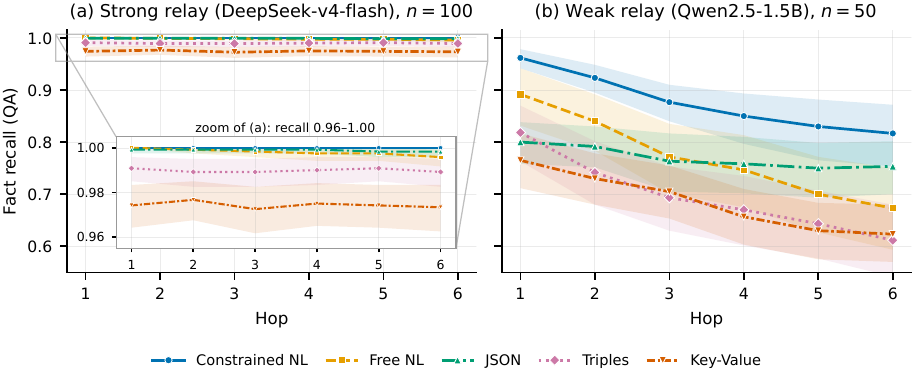}
\caption{Fact recall (\qarec{}) over six relay hops for five message formats (color, marker, and line style jointly encode the format). Bands are bootstrap 95\% CIs over items ($n{=}100$ strong; $n{=}50$ weak). (a)~A strong relay is nearly lossless in every format (inset: zoomed view). (b)~A 1.5B relay breaks the ceiling: on matched items the across-format SD of hop-6 recall grows $8.7\times$ (CI $5.3$--$15.5$); as a range, $2.3\to20.5$ points. \json{} (green triangles) is the only format whose curve stays nearly flat after the first hop. Panels share the y-axis (the inset has its own zoomed scale).}
\label{fig:decay}
\end{figure*}

Figure~\ref{fig:decay}a shows the strong-tier grid.
After six hops every format retains \qarec{} $\geq 0.973$; the multi-stage collapse documented for delegation and translation chains \cite{ao2026reliability,mohamed2025telephone} does not materialize at this tier under faithful-relay instructions.
We read this as a boundary condition: those collapses arise in pipelines whose stages \emph{transform} content, whereas our duty is faithful re-encoding, so collapse is not intrinsic to multi-hop text relay; it requires transformation duty or a weak relay.
What little loss exists is incurred when the brief is first \emph{encoded} into the format, not while the encoded message is relayed.
\kv{}'s curve is flat from hop~1 onward at $\approx0.973$ (a $2.7$-point encode-then-hold deficit), and \json{}/\triples{}/\cnl{} curves are statistically flat throughout (slopes tabulated in Appendix~\ref{app:slopes}).
\fnl{} is the lone exception, drifting slowly but consistently (slope $-0.295$, CI $[-0.379,-0.045]$; raw loss just $0.4$ points over six hops), and the same drift appears under load ($-0.243$, CI $[-0.374,-0.028]$; \S\ref{sec:load}): free prose re-statement drifts where structured and instruction-constrained formats lock.

Two cost observations follow.
First, \cnl{} achieves perfect hop-6 recall ($1.000$) at $143$ tokens per message versus \json{}'s $0.998$ at $194$ tokens: a single precision instruction buys schema-grade fidelity 26\% cheaper in message tokens.
Second, the fidelity--cost trade-off is real but small at this tier: \kv{} is the cheapest format ($135$ tokens) and pays for it with the only measurable content loss.

\subsection{Cognitive load raises cost, not distortion (strong tier)}
\label{sec:load}

\begin{table}[t]
\centering
\small
\setlength{\tabcolsep}{4pt}
\begin{tabular}{@{}lccc@{}}
\toprule
Format & $\Delta$ hop-6 \qarec{} & 95\% CI & Tokens \\
\midrule
\cnl{}    & $+0.000$ & degenerate$^{\dagger}$ & $+45\%$ \\
\fnl{}    & $-0.004$ & $[-0.008,+0.000]$ & $+50\%$ \\
\json{}   & $-0.001$ & $[-0.010,+0.006]$ & $+24\%$ \\
\triples{}& $+0.003$ & $[-0.010,+0.014]$ & $+39\%$ \\
\kv{}     & $-0.001$ & $[-0.011,+0.007]$ & $+53\%$ \\
\bottomrule
\end{tabular}
\caption{Process vs.\ pure relay at the strong tier: paired hop-6 differences ($n{=}60$), bootstrap 95\% CIs, and token change. $^{\dagger}$All 60 paired differences are exactly zero. Every effect is within $\pm 1.4$ points pointwise and $\pm 1.8$ under Bonferroni.}
\label{tab:load}
\end{table}

A natural objection to pure relay is that real agents process while they relay.
The process-relay grid adds an explicit per-hop analysis duty at the strong tier.
Table~\ref{tab:load} shows the result: messages grow 24--53\% longer, yet every format's paired hop-6 difference is small and its CI contains zero.
Rather than merely accepting the null, we read the CIs as equivalence bounds: any load effect on hop-6 fidelity is bounded within $\pm 1.4$ points per format (pointwise), or $\pm 1.8$ points under a Bonferroni-adjusted simultaneous bound across the four non-degenerate formats.
The format ordering and \fnl{}'s chronic drift (slope $-0.243$, CI $[-0.374,-0.028]$) replicate under load.
At this tier, distortion is a property of the \emph{format and the relay model}, not of how busy the relaying agent is.
Whether load interacts with format under weak relays, where fidelity has 20+ points of headroom, is untested here (a ceiling-protected null is cheap; see Limitations).

\subsection{The format effect is tier-dependent}
\label{sec:weak}

\begin{figure}[t]
\centering
\includegraphics[width=0.85\columnwidth]{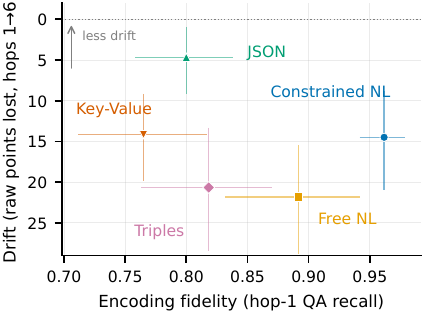}
\caption{The two opposing mechanisms under the weak relay (bootstrap 95\% CIs; $n{=}50$). x-axis: encoding fidelity (hop-1 recall); y-axis: drift as raw points lost over hops 1--6, inverted so up = less drift. Rigid formats pay the largest encoding tolls (\kv{} $0.765$, \json{} $0.800$) but \json{} then drifts least ($4.7$ points); \cnl{} encodes best ($0.962$) yet keeps drifting ($14.5$ points); \fnl{} and \triples{} drift most ($21.8$/$20.7$). The encode--drift trade-off flips the ranking in transit (\json{}: fourth at hop~1, second at hop~6 in 98\% of item-bootstrap resamples).}
\label{fig:mech}
\end{figure}

Figure~\ref{fig:decay}b repeats the pure-relay grid with the weak relay.
The ceiling breaks: every format now decays significantly (all slope CIs exclude zero), and every format degrades relative to the strong tier (paired weak$-$strong hop-6 differences on matched items: $-18$ points for \cnl{} up to $-37$ for \triples{}, all CIs excluding zero).
The across-format dispersion of hop-6 recall grows $8.7\times$ (SD over format means, bootstrap CI $5.3$--$15.5$); as a descriptive range, the spread grows from $2.3$ to $20.5$ points.
Recall at hop~6 runs from $0.817$ (\cnl{}) down to $0.612$ (\triples{}).
The weak model \emph{keeps} format syntax (invalid-\json{} rate 1\%, \S\ref{sec:formats}) while losing content, so the effect is informational, not a parsing artifact.
The judge-free counterpart of the dispersion ratio is $2.2\times$ (CI $1.7$--$3.0$): smaller because the string metric inflates the strong-tier denominator by penalizing legitimate re-rendering (\S\ref{sec:measurement}), so attacks on the QA denominator push the true ratio \emph{up}; orderings and tier gaps replicate under both metrics.
Here ``capability'' denotes a bundled tier contrast (model family, scale, quantization, and serving stack all change); corpus, prompts, protocol, and grader are held fixed.

The decomposition in Figure~\ref{fig:mech} explains how the tier governs the effect: format acts through two mechanisms with opposite signs.
\textbf{Encoding fidelity}: converting a prose brief into a rigid scheme is itself hard for a weak model; the two rigid formats pay statistically indistinguishable hop-1 tolls (\kv{} $0.765$, \json{} $0.800$; paired $\Delta$ CI $[-0.03,+0.10]$), both far below \cnl{} ($0.962$; \cnl{}$-$\json{} $\Delta=0.162$, CI $[0.120,0.205]$).
\textbf{Drift rate}: once encoded, the fixed-key schema anchors content; \json{} drifts least by a wide margin (per-item paired drift advantage: $9.5$ points over \kv{}, $9.8$ over \cnl{}, $16.0$ over \triples{}, $17.2$ over \fnl{}; all four CIs exclude zero, and all four still do under a Bonferroni-4 adjustment, the \kv{} contrast marginally), losing $4.7$ raw points over hops 1--6 while \fnl{} and \triples{} lose $21.8$ and $20.7$ (\cnl{} $14.5$; \kv{} $14.2$).
Two checks rule out survivor bias: first, \kv{} is a toll-matched control, paying the same hop-1 toll as \json{} yet drifting three times as much; second, a survivor-conditioned measure (among facts correct at hop~1, the fraction still correct at hop~6) gives \json{} $0.94$ (CI $0.89$--$0.98$) versus $0.74$--$0.85$ for every other format, with all paired contrasts excluding zero.
Drift resistance is \emph{not} a property of structure per se: line-oriented \triples{} drifts like free prose, and \kv{} like constrained prose.
The pattern is consistent with the \emph{closed, fixed key set} giving every fact a stable address that survives re-encoding (\kv{}'s open per-fact keys lack this), though with one fixed-schema format instantiated we cannot separate the closed key set from nesting or from pretraining familiarity with JSON.
The two mechanisms flip the ranking in transit: \json{} starts fourth of five at hop~1 and finishes second at hop~6 in 98\% of item-bootstrap resamples (\json{}$-$\triples{} at hop~6: $+0.142$, CI $[0.055,0.222]$); the bottom pair (\triples{}, \kv{}) is statistically indistinguishable ($-0.012$, CI $[-0.090,+0.065]$).
Under a strong relay both mechanisms are saturated, which is exactly why the format choice looks free at the frontier and decisive below it.

\subsection{Causal adjudication: faithful, not corrective}
\label{sec:causal}

\begin{figure*}[t]
\centering
\includegraphics[width=\textwidth]{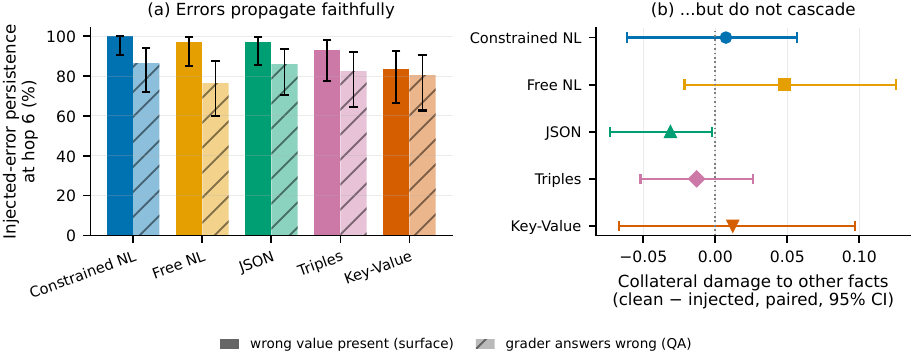}
\caption{Paired-fork injection under the weak relay, conditional on successful injection (target name present at hop 3; per-format $n$: \cnl{} 37, \fnl{} 34, \json{} 35, \triples{} 28, \kv{} 30). (a)~Persistence of the injected error at hop~6: solid bars = wrong value present in the message (surface); hatched bars = grader answers the wrong value (QA); whiskers are Wilson 95\% CIs. Once present, the error survives in 83--100\% of chains in every format. (b)~Collateral damage to the eleven uninjected facts (clean$-$injected, paired bootstrap 95\% CI): no format shows a detectable cascade; \json{} is the only nominally significant cell (1 of 5 uncorrected comparisons) and its sign is mildly protective.}
\label{fig:inject}
\end{figure*}

The drift results invite a tempting reading: rigid schemas act as error-correcting codes for relayed content.
That phrase hides two distinct claims, \emph{resisting change} and \emph{repairing corruption}, and the paired fork separates them.
One honesty note: a wrong owner name is a \emph{redundancy-free leaf error}; no other fact entails the correct value, so no channel could ``correct'' it from in-message evidence alone.
The informative questions are therefore (a) whether formats differ in how faithfully they carry a present error, and (b) whether one corrupted value drags down its neighbors.

\textbf{Persistence is near-universal and direction-blind.}
Once injected, the wrong name survives all three subsequent hops in 83--100\% of chains depending on format (\json{} 97\%, Wilson CI 85--99; \cnl{} 100\%, CI 91--100; \kv{} lowest at 83\%, CI 66--93; Fig.~\ref{fig:inject}a).
At the grader level persistence is 76--86\%.
Because injection conditions on a format-dependent event (the name surviving to hop~3), cross-format comparisons select different item subsets; restricting to the 17 items injectable in \emph{all five} formats, persistence is 82--100\%, so the near-universality is not a selection artifact, though precision is limited there.
Faithfulness is direction-blind: in the clean arms of the same conditioned chains, the \emph{true} name survives to hop~6 at 86--100\% by format, within a few points of the wrong name's 83--100\% format by format.
The lossiest format (\kv{}) is also the one most likely to lose the error.
Where format fidelity \emph{does} matter is upstream, in whether the target fact was still intact at hop~3 to be corrupted at all.

\textbf{No detectable cascade.}
Across all five formats, injecting one wrong value causes no detectable collateral loss on the other eleven facts (Fig.~\ref{fig:inject}b): every paired CI either contains zero or, in the single nominally significant cell (\json{}, $-0.031$, CI $[-0.075,-0.003]$), shows a mildly \emph{protective} sign.
With five uncorrected comparisons, we read that cell as consistent with multiplicity noise; the robust statement is a set of per-format upper bounds on damage (pointwise 95\%): \json{} $\leq 0$, \triples{} $\leq 3$ points, \cnl{} $\leq 6$, \kv{} $\leq 10$, \fnl{} $\leq 13$.
On the 17-item all-format core the estimates are direction-consistent (all CIs cross zero).
Within these bounds, the feared error cascade \cite{xie2026spark} does not ignite from a single corrupted leaf value in any format.

The folklore is thus adjudicated: ``schemas prevent drift'' is true in the narrow sense that rigid structure resists \emph{change of any kind}.
It neither amplifies nor repairs errors; it preserves whatever it is given.
The correct mental model is a faithful, error-localizing channel, not an error-correcting code.

\section{Analysis and Practical Guidance}
\label{sec:discussion}

\paragraph{One phenomenon, four results.}
The four grids compose a single picture.
Format effects on relay fidelity operate through encoding cost and drift resistance; both mechanisms vary with relay tier; at the strong tier both saturate, under weakness both bind; workload moves neither at the tier where this was testable; and the drift resistance that the fixed-key schema buys is direction-blind.
This reconciles the single-hop contradiction in prior work: anti-structure results \cite{tam2024speakfreely,johnson2025nltools} measure the encoding mechanism, pro-structure results \cite{chen2024autoform} measure cost at saturated fidelity, and only the relay regime exposes the drift-side mechanism that rewards structure.

\paragraph{Guidance.}
(1)~\emph{Choose the format for the weakest relay in the pipeline}, not the strongest: at the frontier the choice is nearly free, so the weak link sets the requirement.
(2)~\emph{Short chains or strong relays: precision-instructed NL.} It matches schema fidelity at 26\% fewer message tokens, and the gap widens to ${\sim}63\%$ on billed completion tokens, because the strong relay spends $4\times$ more hidden reasoning tokens emitting \json{} than \cnl{} (Appendix~\ref{app:pareto}), with no schema engineering needed.
(3)~\emph{Long chains through weak relays: a fixed-key schema (\json{}), and budget for the entry toll.} \json{} pays $\sim$$0.16$ recall relative to \cnl{} at encoding but loses only $0.05$ over five further hops.
Within the measured six hops \cnl{} stays ahead on point estimates (hop-6 \cnl{}$-$\json{}: $+0.063$, CI $[-0.010,+0.130]$); the implied crossover near hops 9--10 is an out-of-range extrapolation of the measured loss rates, which we flag as such.
A hybrid design, encode once with a strong model and relay in schema through weak links, inherits both advantages; we flag it as a prediction of the mechanism decomposition, not a tested configuration.
(4)~\emph{Do not expect structure to repair anything}: validation and re-grounding \cite{fan2026martingale} remain necessary; structure only keeps damage local.

\paragraph{Relation to theory.}
Per-format decay parameters give empirical content to relay-fidelity theory: martingale-style accumulation bounds \cite{fan2026martingale} and information-loss limits \cite{ao2026reliability} are format-agnostic; our slopes ($-0.060$ for \json{} to $-0.331$ for \cnl{}, logit/hop under weakness) show the constants those theories abstract over vary several-fold with a design choice as mundane as the message format.

\section{Conclusion}
We completed the missing format~$\times$~hops~$\times$~capability cell in the study of LLM information relay, with programmatic ground truth and a causal injection.
When faithful re-encoding is the instructed objective, the telephone game is not a law of LLM pipelines: it is tier-dependent, absent at our strong tier and nearly an order of magnitude larger in across-format dispersion ($8.7\times$ by QA; $2.2\times$ by the judge-free metric) under our weak relay; we read this as a capability regime, with the bundled-contrast caveat stated throughout.
What the fixed-key schema provides is faithful, error-localizing transmission rather than error correction.
As pipelines push cost-driven small models into relay positions, message format stops being a style choice and becomes a reliability parameter, one that agent infrastructure should expose, log, and tune deliberately.

\section*{Limitations}
\textbf{Synthetic, templated corpus.} Programmatic ground truth buys exact measurement at the price of distributional realism; briefs are information-dense, English-only, and template-rendered with a fixed 12-fact density.
\textbf{Two capability points, bundled tier contrast.} ``Capability governs'' rests on one strong API model versus one 1.5B model that differ jointly in family, scale, quantization, and serving stack.
\textbf{Preservation is the instructed objective.} Every hop is told to re-issue the \emph{complete} handoff; pipelines whose handoffs intentionally compress or transform content face a different objective.
\textbf{The load condition was run only at the strong tier}, where ceiling effects protect the null; a weak$\times$load cell remains untested.
\textbf{Single-leaf injection bounds cascades from below, and correction from above.} We injected an independent fact (an owner name), which is uncorrectable by construction and minimally entangled; errors in facts with dependents could cascade where leaf values do not, and redundancy-bearing errors that a format \emph{could} expose and repair are untested, so ``not corrective'' is established only for the redundancy-free class.
\textbf{Grader reuse and other constants.} The strong grader also serves as the strong relay (separate calls at $t{=}0$ with programmatic exact-match scoring); a fully independent grader family would remove residual self-preference concerns; our judge-free string metric replicates the structured-vs-free separation and the \kv{} deficit (Appendix~\ref{app:mimo}), though near the ceiling its fine ordering differs within noise.
Relay temperature ($0.7$) and chain length (6) were fixed; each item$\times$format chain is a single stochastic realization, so item-level CIs cover item and sampling variance jointly.
QA recall was not calibrated on the pristine hop-0 brief (the hop-0 self-check is string-based), so hop-1 ``encoding tolls'' are upper bounds that may include format-specific grader reading loss; the judge-free replication of the tolls bounds this concern.

\section*{Ethics Statement}
The corpus is fully synthetic; no human subjects or personal data are involved.
The findings inform reliability engineering of multi-agent systems; we do not foresee direct dual-use risk beyond general agent-pipeline capability improvements.

\bibliographystyle{ACM-Reference-Format}
\bibliography{refs}

\appendix
\raggedbottom

\section{Corpus Item and Format Renderings}
\label{app:item}
A software-domain brief contains facts such as: ticket owner, reviewer, on-call contact (person--role); deadline in days, old/new port, canary percentage, pinned version, budget (quantities); migration-order dependency; deploy-window constraint; and a schema-rollback prohibition.
The hop-0 self-check requires every fact's surface form verbatim in the brief (string recall $=1.0$ on all items in all grids).
Format specifications are reproduced verbatim in the released code (\texttt{fmtrelay/formats.py}).

\section{Grader Protocol and Matching Rules}
\label{app:grader}
The grader receives only the hop message and twelve questions, must answer in JSON, and is instructed to output \textsc{Unknown} when the document lacks the answer.
Matching normalizes Unicode, case, and number formatting; pure numbers must match as whole tokens (``17'' does not match ``170''); percentages match on the numeric part; yes/no answers are normalized over a small synonym set.
Parse failures trigger one retry and are otherwise excluded from \qarec{} (rate $0.0\%$ in all grids, including the weak grid).

\section{Cross-Model Replication (String Metric)}
\label{app:mimo}
{\emergencystretch=3em
A partial strong-tier replication on a second strong model (mimo-v2.5-pro) covers 2{,}841 of 3{,}000 relay messages; the run was interrupted by an API budget pause mid-grid, leaving 66--76 complete chains per format.
By the judge-free string metric (hop-6 values computed on complete chains only), hops 1$\to$6: \cnl{} $0.999{\to}0.995$, \json{} $0.999{\to}1.000$, \triples{} $0.987{\to}0.988$, \fnl{} $0.964{\to}0.951$ (the only sustained decay), \kv{} $0.906{\to}0.907$ (encode-then-hold).
The replication reproduces the qualitative structure (encode-step loss, \fnl{} drift, the \kv{} deficit) while the near-ceiling ordering of \cnl{}/\json{} differs within noise.
For the weak grid, the same judge-free metric gives hops 1$\to$6: \cnl{} $0.947{\to}0.815$, \fnl{} $0.873{\to}0.707$, \json{} $0.842{\to}0.800$, \triples{} $0.830{\to}0.645$, \kv{} $0.725{\to}0.593$ (hop-6 spread $22.2$ points), reproducing the top-three QA ordering and the flat-\json{} signature without any LLM judge; the bottom pair swaps relative to QA (\kv{}/\triples{}), consistent with their statistical indistinguishability (\S\ref{sec:weak}).
For reference, the matched-50-item strong-grid hop-6 QA means behind the dispersion ratio are: \cnl{} $1.000$, \fnl{} $0.998$, \json{} $0.997$, \triples{} $0.985$, \kv{} $0.977$.
\par}

\section{Token--Fidelity Trade-off}
\label{app:pareto}
Figure~\ref{fig:pareto} plots the cost--fidelity plane for both capability tiers.
\par\medskip
\noindent\begin{minipage}{\columnwidth}
\centering
\includegraphics[width=\columnwidth]{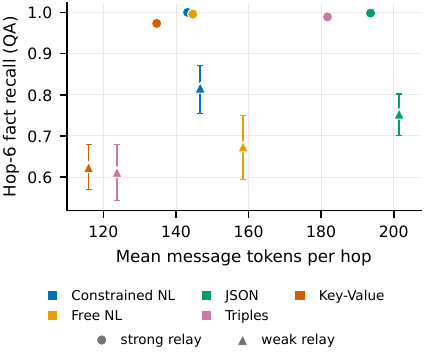}
\captionof{figure}{Hop-6 \qarec{} (bootstrap 95\% CIs) vs.\ mean message tokens per hop (hidden reasoning tokens excluded), both relay tiers (circles: strong; triangles: weak; colors as in Fig.~\ref{fig:decay}). At the strong tier formats trade only cost; under weakness they trade both. Hidden reasoning tokens differ sharply by format at the strong tier (means per hop: \json{} 574, \triples{} 355, \kv{} 329, \cnl{} 139, \fnl{} 86), so billed completion cost favors the NL formats even more than message-token cost does.}
\label{fig:pareto}
\end{minipage}

\section{Per-Format Decay Slopes}
\label{app:slopes}
Table~\ref{tab:slopes} reports the fitted decay slopes behind the drift comparisons of \S\ref{sec:strong} and \S\ref{sec:weak}.
\par\medskip
\noindent\begin{minipage}{\columnwidth}
\centering
\footnotesize
\setlength{\tabcolsep}{3pt}
\begin{tabular}{@{}l>{\raggedright\arraybackslash}p{2.8cm}>{\raggedright\arraybackslash}p{2.8cm}@{}}
\toprule
Format & Strong slope [CI] & Weak slope [CI] \\
\midrule
\cnl{} & $-0.000$ [$-0.000,-0.000$] & $-0.331$ [$-0.459,-0.212$] \\
\fnl{} & $-0.295$ [$-0.379,-0.045$] & $-0.271$ [$-0.361,-0.204$] \\
\json{} & $-0.117$ [$-0.216,0.000$] & $-0.060$ [$-0.115,-0.014$] \\
\triples{} & $-0.007$ [$-0.042,0.021$] & $-0.193$ [$-0.282,-0.120$] \\
\kv{} & $-0.011$ [$-0.045,0.023$] & $-0.143$ [$-0.202,-0.087$] \\
\bottomrule
\end{tabular}
\captionof{table}{Logit-slope estimates per format and tier (bootstrap 95\% CIs). Raw hop-1$\to$6 losses are in \S\ref{sec:strong} and \S\ref{sec:weak}.}
\label{tab:slopes}
\end{minipage}

\section{Injection Example}
\label{app:inject}
For item \texttt{sw-000}, the hop-3 message's owner ``Felix Okafor'' is replaced by ``Hassan Petrov'' (a name absent from the item); the clean and injected arms then relay hops 4--6 independently.
Per-arm, per-hop messages and scores are released with the code.
Model specifics: strong relay/grader DeepSeek-v4-flash (API, accessed June 2026; $t{=}0.7$ relay, $t{=}0$ grading); replication model mimo-v2.5-pro (API, June 2026); weak relay Qwen2.5-1.5B-Instruct, Q4\_K\_M quantization, served via Ollama 0.30.7 on a single RTX 3060.

\end{document}